\pdfoutput=1

\documentclass{article}

\usepackage[final]{neurips_2022}

\usepackage[utf8]{inputenc} %
\usepackage[T1]{fontenc}    %
\usepackage{hyperref}       %
\usepackage{url}            %
\usepackage{booktabs}       %
\usepackage{amsfonts}       %
\usepackage{nicefrac}       %
\usepackage{microtype}      %
\usepackage{xcolor}         %

\usepackage{amssymb}
\usepackage{graphicx}

\usepackage{tabularx}

\usepackage{multirow}
\usepackage{booktabs}
\usepackage{colortbl}  %
\definecolor{Gray}{gray}{0.9}

\definecolor{bittersweet}{rgb}{1.0, 0.44, 0.37}

\usepackage{subcaption}

\usepackage{pifont}%

\usepackage{array}
\newcolumntype{P}[1]{>{\centering\arraybackslash}p{#1}}

\usepackage{amsmath}

\title{Zero-shot Video Moment Retrieval \\ With Off-the-Shelf Models}

\author{%
  Anuj Diwan*, Puyuan Peng*, Raymond J. Mooney \\
  Department of Computer Science, The University of Texas at Austin \\
  \texttt{\{anuj.diwan,pyp\}@utexas.edu, mooney@cs.utexas.edu}
}

\begin{document}

\maketitle

\begin{abstract}
For the majority of the machine learning community, the expensive nature of collecting high-quality human-annotated data and the inability to efficiently finetune very large state-of-the-art pretrained models on limited compute are major bottlenecks for building models for new tasks. We propose a zero-shot simple approach for one such task, Video Moment Retrieval (VMR), that does not perform any additional finetuning and simply repurposes off-the-shelf models trained on other tasks. Our three-step approach consists of moment proposal, moment-query matching and postprocessing, all using only off-the-shelf models. On the QVHighlights~\citep{qvhighlights2021} benchmark for VMR, we vastly improve performance of previous zero-shot approaches by at least 2.5x on all metrics and reduce the gap between zero-shot and state-of-the-art supervised by over 74\%. Further, we also show that our zero-shot approach beats non-pretrained supervised models on the Recall metrics and comes very close on mAP metrics; and that it also performs better than the best pretrained supervised model on shorter moments. Finally, we ablate and analyze our results and propose interesting future directions.
\end{abstract}
{\let\thefootnote\relax \footnotetext{*Co-first authors contributed equally. Order was determined by a coin flip.} }
\section{Introduction}
Given a video and a natural language query, the task of Video Moment Retrieval (VMR) involves temporally localizing moments (i.e. video segments) within the given video that are relevant to the query~\citep{didemo2017,cal,qvhighlights2021}. This task has practical applications in many domains; for example, it can be used for efficiently skipping to parts of the video that are most relevant; this is especially useful with long videos (sports events, television/news broadcasts, and so on). 

It is becoming increasingly clear that data collection is a major bottleneck for progress in machine learning~\citep{https://doi.org/10.48550/arxiv.1811.03402}; it is quite expensive to collect high-quality human-annotated data for a given task of interest. Furthermore, even assuming the existence of high-quality downstream task data, the ability of machine learning practitioners to finetune the best pretrained models on this data is dwindling due to the prohibitively large size of state-of-the-art pretrained Transformer models like CLIP~\citep{clip2021}, GPT-3~\citep{gpt3} and PaLM~\citep{https://doi.org/10.48550/arxiv.2204.02311}; additionally it is often impossible to finetune models if their weights are locked behind an API (e.g. for GPT-3).

We choose the task of Video Moment Retrieval as our task of interest for this case study. This task is related to other tasks like Video Captioning~\citep{xu2016msr-vtt}, Dense Video Captioning~\citep{wang2021end} and Image-Text Matching~\citep{clip2021} in the field of vision-language research and although data and models from these tasks can benefit VMR, there has been no prior work, to our knowledge, on using these models to solve VMR. We ask whether it is possible to bypass these overarching bottlenecks by a) developing \textbf{Zero-shot} methods that require no downstream VMR task data at all and require no finetuning, only inference; and b) using only \textbf{Off-the-shelf} models whose weights are publicly available.

In this work, we explore a three-step approach: a) generating moment proposals, b) performing moment-query matching and c) post-processing, all using off-the-shelf models in a novel zero-shot manner. In particular, we use a simple shot transition detector~\citep{pyscenedetect} for the first step, an off-the-shelf CLIP~\citep{clip2021} model and a Video Captioning model (UniVL: ~\cite{univl2020}) for the second step, and post-processing using a simplified version of the Watershed~\citep{10.5555/2372488.2372495} algorithm. Our overall approach is depicted in Figure~\ref{fig:flow}. We evaluate on the QVHighlights~\citep{qvhighlights2021} benchmark and show that using our simple pipeline, we vastly improve performance on a previous zero-shot approach (between 2.5x-5x on all metrics), and surprisingly beat the best non-pretrained models that are explicitly trained on VMR QVHighlights data on the Recall metrics while coming very close on the mAP metrics. Although our approach still lags behind the current state-of-the-art model on QVHighlights, we hope that our proposed future improvements described in Section~\ref{sec:future} can help bridge the gap; we conduct an oracle experiment to check how far we can potentially go with shot-detection-based moment proposals. In addition, we run a length-wise analysis of performance that shows that while all models (including ours) struggle to perform well on short moments, our models perform much better than state-of-the-art models on short moments. 

In summary, our contributions include: a) proposing novel zero-shot approaches for Video Moment Retrieval that only use off-the-shelf public models b) demonstrating that zero-shot approaches are extremely strong baselines for VMR and can even beat models supervised on in-domain task data on some metrics and c) showing a length-based discrepancy in performance and improving performance for short moments beyond the best approach. 

\section{Related Work}
\label{sec:related}

\textbf{Video Moment Retrieval: Datasets.} Datasets for VMR include DiDeMo~\citep{didemo2017}, Charades-STA~\citep{tall2017}, ActivityNet Captions~\citep{activitynetcaptions2017}, TVR~\citep{tvr2020} and QVHighlights~\citep{qvhighlights2021}. Each example in most of these datasets consists of a video, a natural language query, and ground truth moment timestamps (multiple for QVHighlights and one for all other datasets) for localizing the query. QVHighlights also provides additional \textit{saliency scores} that are frame-level annotations on a five-point Likert scale for how well the frame matches the query.

\textbf{Video Moment Retrieval: Approaches.}
Recent state-of-the-art approaches for Video Moment Retrieval including Moment-DETR~\citep{qvhighlights2021} and UMT~\citep{umt}) involve designing an end-to-end deep neural network that produces segment proposals and confidence scores simultaneously and training this network end-to-end using a moment retrieval dataset. Older approaches like MCN~\citep{didemo2017}, CAL~\citep{cal} and XML~\citep{tvr2020} are not end-to-end; they consist of a multi-step pipeline that usually consists of generating moment proposals followed by either regressing/predicting moment indices or scoring moment proposals. Often these methods also have a post-processing non-maximal suppression step. Each component is trained separately using a moment retrieval dataset.

\textbf{Zero-shot transfer with CLIP.} The rise of large pretrained models trained on large quantities of data like CLIP~\citep{clip2021}, BERT~\citep{bert}, GPT-3~\citep{gpt3} and other so-called Foundation Models has unlocked the ability to transfer knowledge to new tasks without either ever training on the new task (`zero-shot') or with very few examples of the task (`few-shot'). CLIP, in particular, has been used in innovative zero-shot ways for out-of-distribution detection~\citep{https://doi.org/10.48550/arxiv.2109.02748}, text-to-shape generation~\citep{https://doi.org/10.48550/arxiv.2110.02624} and referring-expression comprehension~\citep{subramanian2022reclip} in addition to zero-shot image classification as studied in the original CLIP paper~\citep{clip2021}.

\section{Video Moment Retrieval}
\begin{figure*}[!h]
    \centering
    \includegraphics[width=1\textwidth]{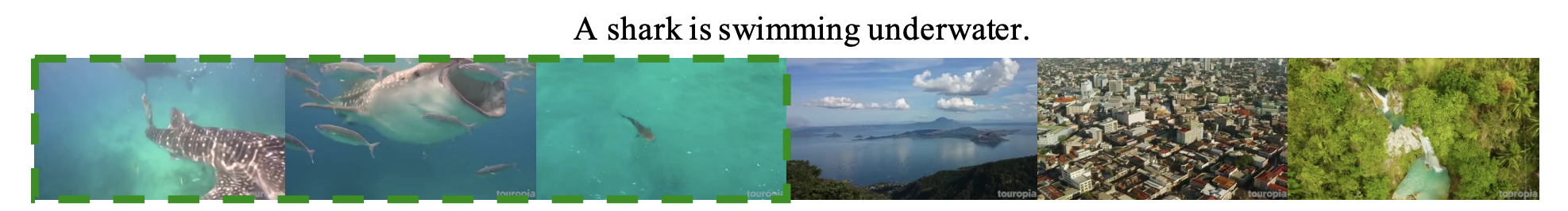}
    \caption{Moment retrieval example from QVHighlights~\citep{qvhighlights2021}}
    \label{fig:vmreg}
\end{figure*}

The task of Video Moment Retrieval (VMR) involves temporally localizing moments (i.e. video segments) within a video given a natural language query. Formally, given a video $V$ and a query $Q$, a VMR model must be able to output $K$ segments $v_1, \ldots, v_K$ (where each segment $v_i$ consists of a start and end timestamp $\text{start}_i$ and $\text{end}_i$) with corresponding scores $s_1, \ldots, s_K$ that denote how similar the model believes the segment is to the query. We describe commonly used datasets and approaches for this task in Section~\ref{sec:related} and the evaluation metrics in Section~\ref{subsec:datasets}. Figure~\ref{fig:vmreg} shows an example from a recently released dataset for this task, QVHighlights~\citep{qvhighlights2021}; provided the query \texttt{`A shark is swimming underwater'} and a nature documentary video, the model must be able to output the predicted relevant video moments and the confidence score.
\section{Our Approach}
\label{sec:approach}

\begin{figure*}[h]
  \centering
      \includegraphics[width=\textwidth]{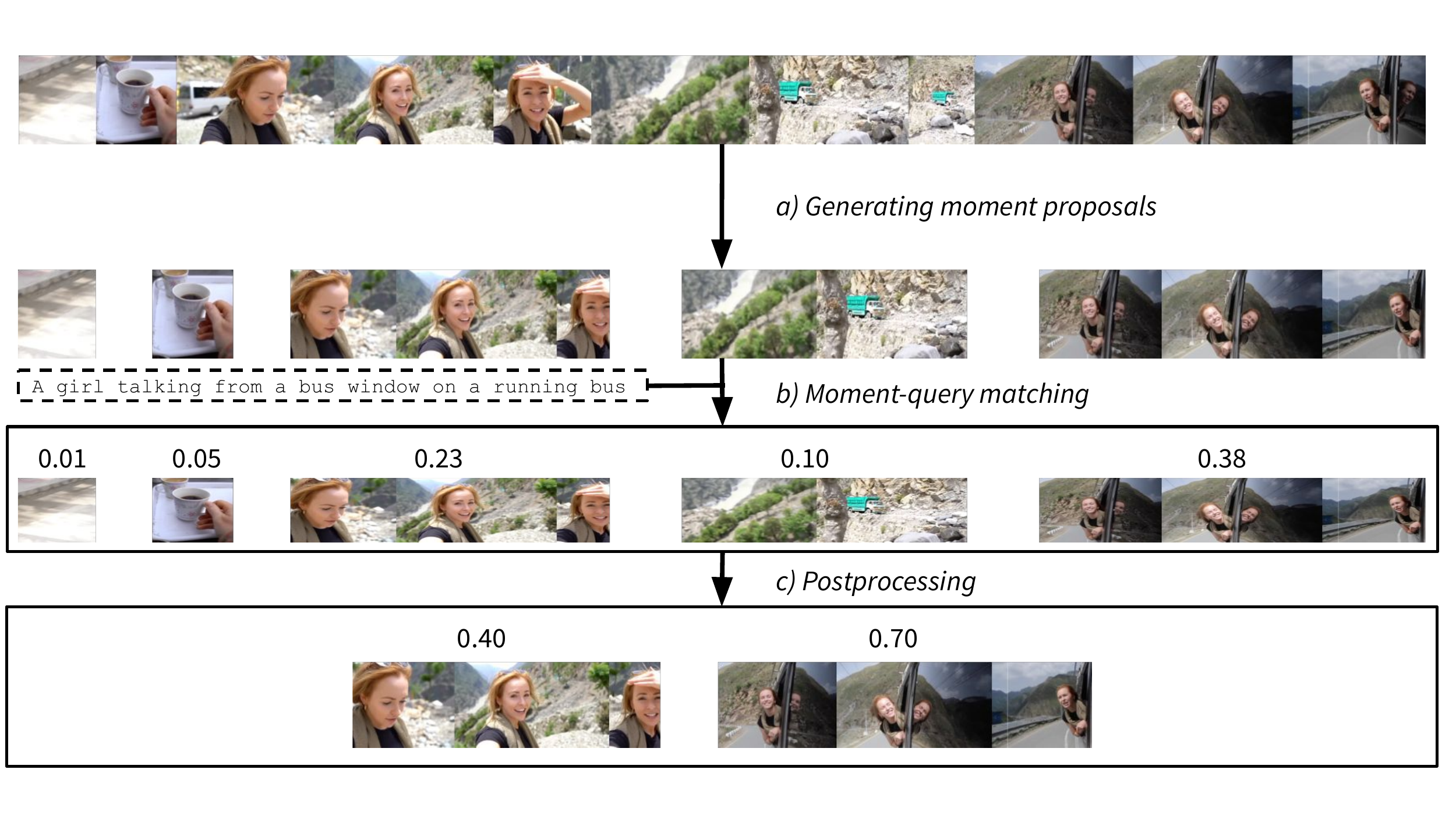} 
      \caption{Our overall three-step procedure.}
      \label{fig:flow}
\end{figure*}

As discussed in Section~\ref{sec:related}, state-of-the-art VMR models devise an end-to-end neural network to produce proposals and scores simultaneously while older methods separately train proposal and scoring models. In contrast, we are interested in developing an \textbf{untuned} method for VMR i.e. a \textbf{zero-shot} method that does not require any VMR training data at all. In addition, we would like to build it using publicly available \textbf{off-the-shelf} models. With that spirit, we develop a three-step approach consisting of a) generating moment proposals b) performing moment-query matching and c) postprocessing. This overall procedure is depicted in Figure.~\ref{fig:flow}. This closely resembles older VMR approaches discussed in Section~\ref{sec:related}, but with the important distinction that unlike those approaches we use off-the-shelf models for each step without any further training required.

\subsection{Generating Moment Proposals.}
Our first step is generating segment or moment proposals from the provided video i.e. given the video $V$ of length $T$ seconds and a natural language query $Q$, we generate $K$ moment proposals with ${v_1, v_2, \ldots, v_K}$ with lengths ${t_1, t_2, \ldots, t_K}$. To do this, we propose a simple method \textbf{ShotDetect}; using preexisting shot transition detection~\footnote{\url{https://en.wikipedia.org/wiki/Shot_transition_detection}} software~\citep{pyscenedetect}, we obtain several disjoint segments of the video, where, informally, each segment consists of contiguous frames with smooth camera motion. There is a rich body of literature on improving shot transition detection (a few recent papers include \citet{7040826}, \citet{7079058}, \citet{https://doi.org/10.48550/arxiv.2008.04838}, \citet{https://doi.org/10.48550/arxiv.1705.03281} and \citet{DBLP:journals/sivp/SinghTC20}) and we choose to use the content-aware detector from the open-source PySceneDetect~\footnote{\url{http://scenedetect.com/en/latest/}} toolkit which has been used for other video understanding tasks as well~\citep{Wu_2021_CVPR}. This detector works by triggering scene changes when adjacent frames have a significant color change that is greater than a provided threshold $\lambda$; we treat $\lambda$ as a hyperparameter.

In order to ablate the choice of moment proposal method, we also devise a baseline moment proposal method called \textbf{SlidingWindow}; here, we simply generate $15$-second long proposal windows with a stride of $10$ seconds.

\subsection{Moment-Query matching}
In this step, we would like to score every moment proposal using a similarity score $s \in [0.0,1.0]$ i.e. a similarity score $s_k$ between every segment $v_k$ and the provided query $Q$. We explore two different methods, \textbf{VideoCaptioning} and \textbf{CLIP} to obtain this similarity score.

\begin{figure*}[h]
  \centering
    \begin{subfigure}{.49\linewidth}
      \includegraphics[width=\textwidth]{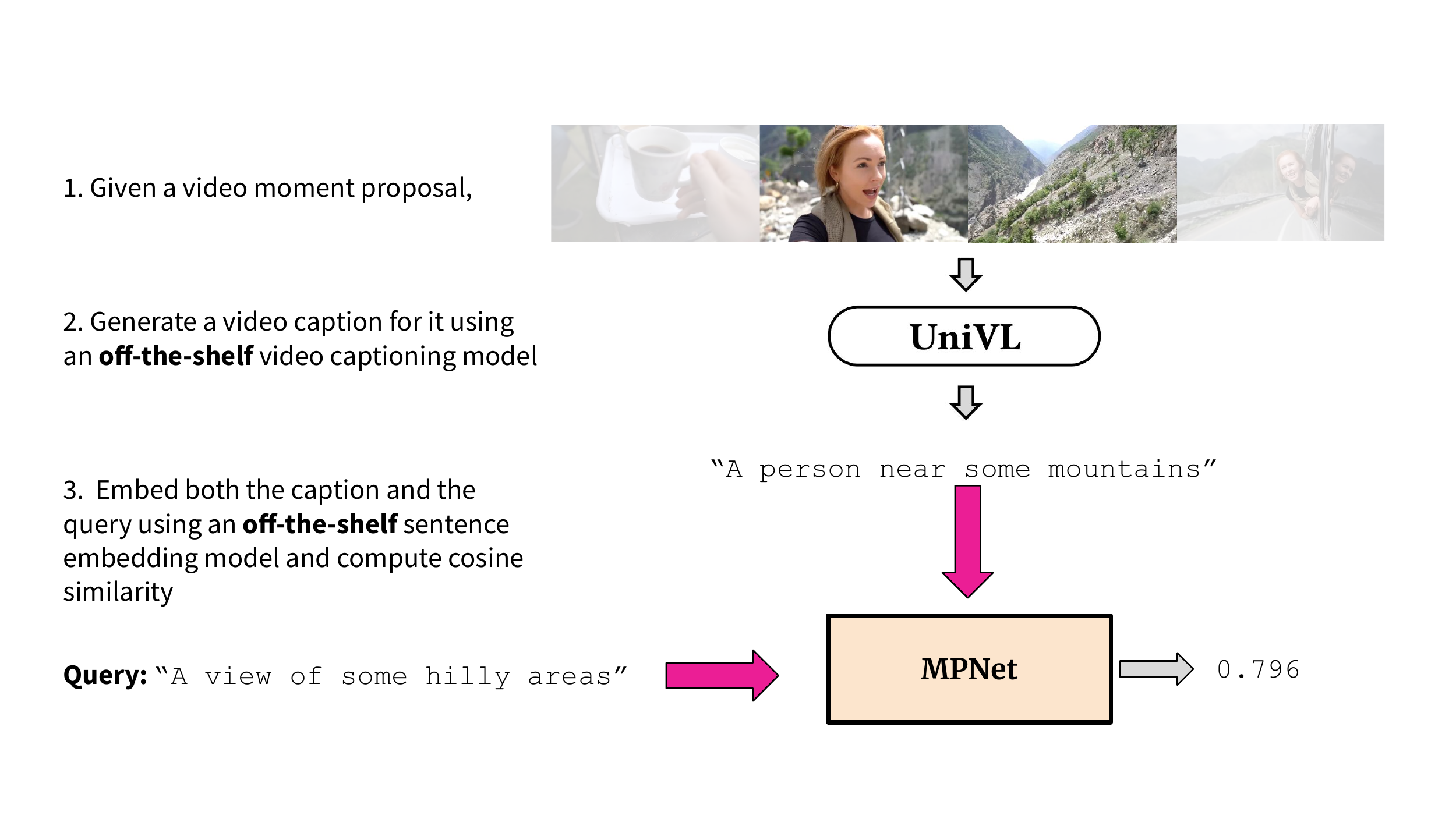} 
      \caption{VideoCaptioning}\label{fig:vidcap}
      \end{subfigure} %
      \begin{subfigure}{.49\linewidth}
      \includegraphics[width=\textwidth]{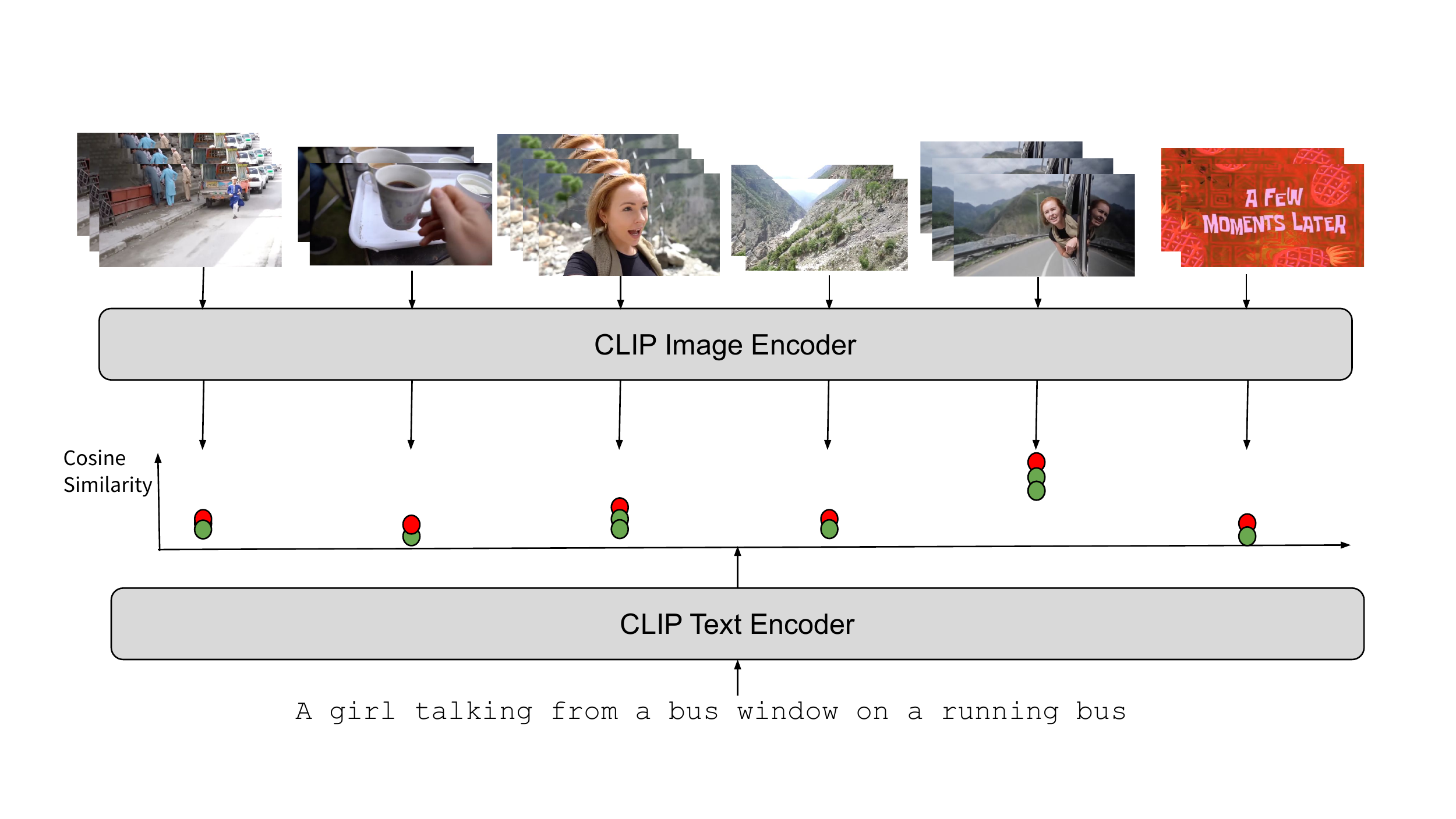}
      \caption{CLIP}\label{fig:clip_embed}
      \end{subfigure}
      \caption{Moment-Query Matching}
\end{figure*}

\noindent \textbf{VideoCaptioning}. We first run an off-the-shelf video captioning model UniVL~\citep{univl2020} on each video segment $\{v_1, v_2, \cdots, v_K\}$ producing natural language captions $\{l_1, l_2, \ldots, l_K\}$. Next, we use an off-the-shelf sentence embedding model MPNet~\citep{https://doi.org/10.48550/arxiv.2004.09297} denoted as $E$ (note that the sentence embedding is obtained by mean-pooling individual token embeddings) to embed both the video segment captions and the natural language query into $\{E(l_1), E(l_2), \ldots, E(l_K)\}$ and $E(Q)$. Finally, we compute the cosine similarity between the query embedding and each caption embedding to obtain the final similarity scores $\{s_1, s_2, \ldots, s_K\}$. Figure~\ref{fig:vidcap} illustrates the similarity score computation with VideoCaptioning.

\textbf{CLIP}. For every video segment $v_k, k \in \{1,\cdots,K\}$ with a length of $t_k$ seconds, we first sample $M_k = \lceil t_k \rceil $ frames from $v_k$ at a rate of $1$ fps; denote these frames as $f^{(k)}_{1},\ldots,f^{(k)}_{M_k}$. We then embed every sampled frame using the CLIP~\citep{clip2021} image encoder. Similarly, we use the CLIP text encoder to embed the query $Q$. Taking the cosine similarity between the embeddings of each frame and query, we get $\{s^{(k)}_1, s^{(k)}_2, \ldots, s^{(k)}_{M_k}\}$. We then calculate the similarity score between segment $v_k$ and query $Q$ as $s_k = f(s^{(k)}_1, s^{(k)}_2, \ldots, s^{(k)}_{M_k})$ where $f$ is an aggregation function. We only consider a simple aggregation function $f$, namely \texttt{max}. Applying this procedure to every video segments, we get similarity scores $\{s_1, s_2, \cdots, s_K\}$. Figure.~\ref{fig:clip_embed} illustrates this process with \texttt{max} as the aggregation function $f$.
\subsection{Post-processing}
We can use the computed video segments and corresponding similarity scores from the previous step as-is without post-processing. Alternatively, we propose a post-processing step called \textbf{Simple Watershed} which is a simplified version of the Watershed algorithm~\citep{10.5555/2372488.2372495} that is a popular technique in image segmentation~\citep{Beucher2009THEWT} and is used to post-process the CLIP baseline for VMR in~\citet{qvhighlights2021}. We use a simplified variant of Watershed as it is easier to describe and work with.

In SimpleWatershed, given a sequence of segments $v_1,\ldots,v_k$ and similarity scores $s_1, \ldots, s_k$ and a similarity score threshold $\gamma$ (which is a hyperparameter), for every consecutive run of segments $v_i,\ldots,v_j$ where $s_k \ge \gamma \ \forall i \le k \le j$, we merge the run into a single segment and set its similarity score to $\max{(s_i,\ldots,s_j)}$.

\section{Experiments}
\subsection{Experimental Setup}
\label{subsec:expsetup}
For Video Captioning, we use the UniVL model~\citep{univl2020} pretrained on HowTo100M~\citep{https://doi.org/10.48550/arxiv.1906.03327} finetuned for video captioning on the MSRVTT~\citep{xu2016msr-vtt} using the UniVL codebase~\footnote{\url{https://github.com/microsoft/UniVL}}.
For CLIP~\citep{clip2021}, we use the \texttt{CLIP-ViT-B-32}\footnote{\url{https://huggingface.co/sentence-transformers/clip-ViT-B-32}} pretrained model. We obtained the best value of $\lambda$, the PySceneDetect color change threshold as $53$ for both moment-query matching methods without post-processing and $\lambda=32, \gamma=0.7$ for SimpleWatershed. We ablate $\lambda$ in Appendix~\ref{sec:applambda}. We run experiments on an RTX 8000 with 45 GB of GPU RAM.

\subsection{Datasets and Evaluation}
\label{subsec:datasets}

We evaluate our model on the QVHighlights~\citep{qvhighlights2021} dataset.
Each example consists of a video, a natural language query, and multiple ground truth moment timestamps for localizing the query. In addition, QVHighlights also provides additional saliency scores that
are frame-level annotations on a five-point Likert
scale for how well the frame matches the query. We note that collecting frame-level annotation data is extremely expensive and not done in previous VMR datasets; therefore, we place models that are trained on this additional data in a different category. We recap the QVHighlights dataset by borrowing figures and statistics from \citet{qvhighlights2021} in Appendix~\ref{sec:appendixqv}.

The QVHighlights dataset has a train-val-test split of $70\% - 15\% - 15\%$. However, the test set annotations are private. Further, while the validation set has $1550$ videos, we were able to obtain only a subset consisting of $1434$ videos; the other videos were either no longer on YouTube or failed to download. Thus, we evaluate our models on this subset that we call the \textit{filtered validation} set or \textbf{val-filt} for short. Table~\ref{tab:comparesets} shows the same model Moment-DETR from~\citet{qvhighlights2021} evaluated on these three sets (test, val, val-filt). We find that all three sets result in approximately the same evaluation metric results and thus could be fairly interchangeable with respect to analysis. However, to ensure completely fair comparison, we report metrics on the val-filt subset (which we have access to) for as many models as possible and the test subset for other models (whose metrics are directly taken from~\citet{qvhighlights2021}) with the understanding that metrics on val-filt and test can be likely fairly compared to each other.

To evaluate our models, we use the same evaluation code and metrics as~\citet{qvhighlights2021}; a) Mean Average Precision (mAP) with IoU thresholds $0.5$ and $0.75$, as well as the average mAP over multiple IoU thresholds and b) Recall@1 with IoU thresholds $0.5$ and $0.7$. Please refer to~\citet{qvhighlights2021} fo a detailed description of these evaluation metrics.

\begin{table*}[h]
    \small
    \centering
    \caption{Comparing results on Moment-DETR w/ PT~\citep{qvhighlights2021} on different evaluation sets. The test and val set results are taken directly from~\citet{qvhighlights2021} while we evaluate ourselves on the val-filt set.}
\begin{tabular}{lccccc}
\toprule
\multirow{2}{*}{ Evaluation Set } & \multicolumn{2}{c}{R1} & \multicolumn{3}{c}{mAP}\\
\cmidrule(l){2-3} \cmidrule(l){4-6} 
& @0.5 & @0.7 & @0.5 & @0.75 & avg \\
\midrule
test & 59.78 & 40.33 & 60.51 & 35.36 & 36.14 \\
val & 59.68 & 40.84 & - & - & 36.30 \\
val-filt & 59.74 & 41.10 & 59.90 & 35.42 & 36.19 \\
\bottomrule
\end{tabular}
    \label{tab:comparesets}
    \end{table*}

\subsection{Baselines}
\label{subsec:baselines}
We compare our model with three types of baselines: a) VMR-Supervised, b) VMR-Supervised + Saliency and c) Zero-Shot approaches.

\noindent \textbf{VMR-Supervised}. These models are trained on the QVHighlights training dataset excluding the saliency score annotations. We select the same baselines as the ones considered in the QVHighlights paper~\citep{qvhighlights2021}; two proposal based models MCN~\citep{didemo2017} and CAL~\citep{cal} and two span prediction models, XML~\citep{tvr2020} and Moment-DETR w/o saliency~\citep{qvhighlights2021}. We reproduced Moment-DETR w/o saliency ourselves since the checkpoint for this setting is not publicly available.

\noindent \textbf{VMR-Supervised + Saliency}. These models are trained on the QVHighlights training set, including saliency score annotations. We compare with XML+, Moment-DETR and Moment-DETR w/ Pretraining (pretrained on ASR Captions) all from~\citet{qvhighlights2021} and UMT~\citep{umt}, the current SoTA on QVHighlights that is trained on an additional audio modality. Note that unlike zero-shot approaches, the architecture, loss and training scheme of these models have been specifically developed for Video Moment Retrieval and have been trained on in-domain supervised data.

\noindent \textbf{Zero-shot}. These models (including ours) have not been explicitly trained on QVHighlights. We compare against the only baseline known to us: the CLIP+Watershed baseline provided in~\citet{qvhighlights2021}. Our CLIP-based approach is different from this baseline. We use segment proposals obtained from shot detection software and provide results with and without SimpleWatershed postprocessing. On the other hand, CLIP+Watershed computes image-text similarities on frames sampled at $0.5$ Hz and then runs the full Watershed algorithm on them; essentially, it just uses 2-second moment proposals instead of obtaining them from shot detectors.

\subsection{Oracular Bounds}
\label{sec:oracularbounds}
Our approach is based on the video segments provided by PySceneDetect, which introduces a possible weak link into our approach: if the segments predicted by this tool are of low quality, it will be impossible to ever succeed on QVHighlights, even given a perfect Moment-Query Matcher. To ensure that this is not the case, we compute oracular bounds to gauge whether high scores are achievable. We design our `perfect moment-query matcher' (which has access to the ground truth moments) to be such that predicted similarity scores nearly perfectly reflect the match between video segments and the query. 

\noindent \textbf{Non-Postprocessing Oracular Bound.}  We define the similarity score between a given segment and the query to be the highest IoU\footnote{There can be oracular similarity scores that are better than IoU and thus our oracular bounds are not upper bounds.} between the segment and the nearest ground truth moment.

\noindent \textbf{Postprocessing Oracular Bound.} We merge adjacent segments (for the SimpleWatershed post-processing step) as long as doing so will increase the IoU of the combined segment with the nearest ground truth moment.

\section{Results}
\label{sec:results}
\subsection{Main Results}
\begin{table*}[t!]
    \setlength{\tabcolsep}{0.4em}
    \small
    \centering
    \caption{Results using all evaluation metrics on the QVHighlights val-filt set (for rows without an asterisk *) or the test set (for rows with an asterisk *). Bold values denote the best-performing approaches within each category.}
\begin{tabular}{llccccc}
\toprule
\multirow{2}{*}{ Category }& \multirow{2}{*}{ Method } & \multicolumn{2}{c}{R1} & \multicolumn{3}{c}{mAP}\\
\cmidrule(l){3-4} \cmidrule(l){5-7} 
& & @0.5 & @0.7 & @0.5 & @0.75 & avg \\
\midrule
&CLIP+Watershed(~\citep{qvhighlights2021}) & 16.88 & 5.19 & 18.11 & 7.00 & 7.67 \\
& SlidingWindow + VideoCaptioning \textbf{(Ours)} & 19.60 & 6.00 & 25.94 & 6.00 & 9.58 \\
&ShotDetect+VideoCaptioning \textbf{(Ours)} & 22.25&14.71&28.90&17.30&18.06 \\
& SlidingWindow + CLIP \textbf{(Ours)} & 29.71 & 8.86 & 35.26 & 8.31 & 13.42 \\
&ShotDetect+CLIP \textbf{(Ours)} & 40.24 & 25.94 & 41.74 & 24.11 & 24.82 \\ 
\multirow{-6}{*}{Zero Shot}&ShotDetect+CLIP+SimpleWatershed \textbf{(Ours)} & \textbf{48.33} & \textbf{30.96} & \textbf{46.94} & \textbf{25.75} & \textbf{27.96} \\
\midrule
&MCN*(~\cite{didemo2017}) & 11.41 & 2.72 & 24.94 & 8.22 & 10.67 \\
&CAL*(~\cite{cal}) & 25.49 & 11.54 & 23.40 & 7.65 & 9.89 \\
&XML*(~\cite{tvr2020}) & 41.83 & \textbf{30.35} & 44.63 & \textbf{31.73} & \textbf{32.14}  \\
\multirow{-4}{*}{VMR-Sup} &M-DETR w/o saliency loss(~\cite{qvhighlights2021}) & \textbf{45.03} & 25.81 & \textbf{48.42} & 21.91 & 24.68 \\
\midrule
&XML+*(~\cite{qvhighlights2021}) & 46.69 & 33.46 & 47.89 & 34.67 & 34.90 \\
&M-DETR*(~\cite{qvhighlights2021}) & 52.89 & 33.02 & 54.82 & 29.40 & 30.73 \\
&M-DETR w/ PT(~\cite{qvhighlights2021}) & 59.74 & 41.10 & 59.90 & 35.42 & 36.19 \\
&M-DETR w/ PT*(~\cite{qvhighlights2021}) & 59.78 & 40.33 & \textbf{60.51} & 35.36 & 36.14 \\
&UMT*(~\cite{umt}) & 56.23 & 41.18 & 53.83 & 37.01 & 36.12 \\
\multirow{-6}{*}{VMR-Sup+Saliency} &UMT w/ PT*(~\cite{umt}) & \textbf{60.83} & \textbf{43.26} & 57.33 & \textbf{39.12} & \textbf{38.08} \\
\bottomrule
\end{tabular}
    \label{tab:baseline_comparison}
    \end{table*}
We present our main results in Table~\ref{tab:baseline_comparison}. As described in Section~\ref{subsec:baselines}, we categorize the approaches we compare into VMR-Supervised, VMR-Supervised + Saliency and Zero-Shot. Further, as discussed in Section~\ref{subsec:datasets}, we report our approaches and some baselines on the val-filt validation set; for others (marked with an asterisk *) we report on the test set. Note that from Table~\ref{tab:comparesets} and by comparing \texttt{M-DETR w/ PT*} with \texttt{M-DETR w/ PT} in Table~\ref{tab:baseline_comparison} one can conclude that results on the test set and the val-filt set are fairly comparable.

\noindent \textbf{Comparison with Zero-shot.} Comparing zero-shot approaches, we see that all of our approaches significantly outperform the sole zero-shot baseline (CLIP+Watershed) from~\cite{qvhighlights2021}; our best approach, ShotDetect+CLIP+SimpleWatershed raises the zero-shot performance at least by 2.5x on all metrics and by as much as 5x on R1@0.7 without ever having seen data from VMR, or indeed, the video modality (since CLIP was trained solely on images and text). We note that our ShotDetect approach performs significantly better than our SlidingWindow approach (for both VideoCaptioning and CLIP), showing that ShotDetect is indeed helpful for VMR as compared to the baseline. Furthermore, CLIP works much better than UniVL VideoCaptioning; we hypothesize that this is because the generated video captions (trained on MSRVTT~\cite{xu2016msr-vtt}) are not in-domain for QVHighlights queries, which may make matching captions to queries harder.

\noindent \textbf{Comparison with VMR-Supervised.} Surprisingly, our best performing zero-shot approach ShotDetect+CLIP+Watershed outperforms all VMR-Supervised models (MCN, CAL, XML and Moment-DETR w/o saliency loss) on both Recall metrics. It outperforms Moment-DETR w/o saliency loss on four out of five metrics. This shows that a simple approach like ours that uses no supervised data can perform competitively or even beat supervised approaches.  

\noindent \textbf{Comparison with VMR-Supervised + Saliency (+ Pretraining).}
Here we can see that our approach performs worse than the worst method (XML+) in this category and our approach is significantly worse than the best approach, UMT w/ PT. We first note that adding saliency score supervision (an expensive annotation) provides a huge jump in performance (compare M-DETR to M-DETR w/o saliency loss). Although we bridged the gap between zero-shot and normal supervised approaches, there is still a significant gap between zero-shot and pretrained models or supervision with frame-level annotations like saliency scores. In order to bridge this gap, it will be essential to understand how to obtain saliency score-like information in a zero-shot manner to incorporate into our models, and we leave this to future work.

\subsection{Oracular bounds}
\label{subsec:oracularresults}
Table~\ref{tab:upperbound} shows the two oracular bounds introduced in section~\ref{sec:oracularbounds} for our two best zero shot approaches (with and without SimpleWatershed post-processing). The Non-Postprocessing Oracular Bound, which bounds our ShotDetect+CLIP approach, is comparable to the best approaches from Table~\ref{tab:baseline_comparison} in terms of Recall but is much worse at the mAP score, indicating that, without post-processing, we are likely upper bounded by a poor mAP score. The Postprocessing Oracular Bound, which bounds our ShotDetect+CLIP+SimpleWatershed approach outperforms the best supervised UMT w/ PT by a very large margin. However, note that this Oracular Bound will be very hard to attain, even compared to the Non-Preprocessing Oracular Bound, as one needs both perfect similarity scores and segment merging algorithms.

\begin{table*}[t!]
    \centering
    \small
    \caption{Oracular Bounds}
\begin{tabular}{llccccc}
\toprule
\multirow{2}{*}{ Category }& \multirow{2}{*}{ Method } & \multicolumn{2}{c}{R1} & \multicolumn{3}{c}{mAP}\\
\cmidrule(l){3-4} \cmidrule(l){5-7} 
& & @0.5 & @0.7 & @0.5 & @0.75 & avg \\
\midrule
Zero Shot & ShotDetect + CLIP\textbf{(Ours)} & 40.24 & 25.94 & 41.74 & 24.11 & 24.82\\ 
Oracular Bound & Non-Postprocessing Bound ($\lambda = 53$)& 63.94 & 41.49 & 54.29 & 29.41 & 30.98 \\ 
\midrule
Zero Shot&ShotDetect+CLIP+SimpleWatershed\textbf{(Ours)} & 48.33 & 30.96 & 46.94 & 25.75 & 27.96\\
Oracular Bound & Postprocessing Bound  ($\lambda = 47$) & 87.94 & 79.01 & 81.02 & 67.23 & 63.99 \\
\bottomrule
\end{tabular}
\label{tab:upperbound}
\end{table*}

\subsection{Length-wise analysis of performance}\label{sec:length}
Next, we perform a more fine-grained comparison of our approaches with supervised models by evaluating on different subsets of the val-filt set determined by the lengths of the ground truth moments, in order to understand how models perform on short vs. middle vs. long ground truth moments. \textbf{Short} moments have less than $10$ seconds, \textbf{Medium} moments are between $10$ and $30$ seconds and \textbf{Long} moments are greater than $30$ seconds, which are, respectively, 27.63\%, 62.41\% and 36.26\% of the total set of moments. Using the avg. mAP metric as our metric of comparison, we present these results in Table~\ref{tab:detailed}. We first note that for all approaches, the short mAP is very low in absolute terms and these models score well on the overall avg. mAP metric (that ignores length) only due to long and medium moments. This indicates a systematic failure of all approaches on a specific subset of the QVHighlights dataset and should be taken into consideration when building or using models for this task. Our zero-shot approach ShotDetect+CLIP is much better at predicting short moments, while supervised methods are better at predicting long moments. We do not have a good explanation for this phenomenon, but it could be a good hint to build better inductive bias like ours (implicitly from the ShotDetection step) into the design of supervised models. Note that using the SimpleWatershed post-processing still beats the supervised baselines on the short avg mAP, which is encouraging, but has lower performance as compared to without post-processing (4.90 vs 7.08). Instead, the post-processing helps the long (33.19 vs 27.49) and middle (28.44 vs 26.15) mAPs. This makes intuitive sense as the SimpleWatershed algorithm combines segments of high similarities into longer segments.

\begin{table*}[h]
    \small
    \centering
    \caption{Length-wise analysis of performance using the avg. mAP metric}
\begin{tabular}{llccc}
\toprule
Category& Method &Long&Medium&Short \\
\midrule
VMR-Supervised &M-DETR w/o Saliency loss & 31.38&25.29&2.63 \\
\midrule
& M-DETR & 41.11&32.3&3.28 \\
\multirow{-2}{*}{VMR-Supervised w/ Saliency}&M-DETR w/ PT &\textbf{45.18}&\textbf{37.53}&3.50 \\
\midrule
&ShotDetect+CLIP &27.49&26.15&\textbf{7.08} \\
\multirow{-2}{*}{Zero Shot}&ShotDetect+CLIP+SimpleWatershed &33.19&28.44&4.90 \\
\bottomrule
\end{tabular}
    \label{tab:detailed}
    \end{table*}

\subsection{Relaxing the zero-shot assumption}
Although this falls outside our main setting of untuned zero-shot approaches, we also investigate the possibility of fine-tuning CLIP on the QVHighlights training set. This investigation is largely preliminary.

We use the original CLIP image encoder to encode RGB frames of the video, the original CLIP text encoder to encode the query, and use the same CLIP contrastive loss as the training objective. Negative frames are either sampled from a) the set of frames in the same video that have low (original) CLIP similarity with the text query (`hard' negative examples as they are from the same video), b) other videos within the same batch (`easy' negative examples). We tune the following hyperparameters: the number of early CLIP layers to freeze, the number of hard and easy negative examples, the number of positive examples, and the learning rate. We found that freezing early layers during fine-tuning is helpful, and increasing the number of negative examples is also helpful, although we couldn't scale it up as much as CLIP's original training recipe due to computational limitations. However, in all experiments, we did not see a significant improvement over the original zero-shot result, and the validation performance usually starts to decrease after a few hundred steps. In Table~\ref{tab:finetune}, we list our best fine-tuning performance and compare it with zero-shot performance. We obtained this result with the following hyperparameters: freezing the first $8$ layers of both CLIP encoders, $30$ hard examples, and $200$ negative examples, $1$ positive example (i.e. we only sample one frame from the interval that the query corresponds to), with a learning rate schedule involving a linear ramp-up to $10^{-6}$ for $10\%$ steps and a linear decrease to $0$ for the remaining $90\%$ steps.

\begin{table*}[t!]
    \centering
    \small
    \caption{CLIP fine-tuning performance}
\begin{tabular}{llcccc}
\toprule
\multirow{2}{*}{ Category }&  \multicolumn{2}{c}{R1} & \multicolumn{3}{c}{mAP}\\
\cmidrule(l){2-3} \cmidrule(l){4-6} 
& @0.5 & @0.7 & @0.5 & @0.75 & avg \\
\midrule
Zero-shot & 40.24 & 25.94 & 41.74 & 24.11 & 24.82 \\
Fine-tuning & 42.12 & 27.89 & 43.00 & 24.68 & 25.50 \\
\bottomrule
\end{tabular}
\label{tab:finetune}
\end{table*}

\section{Future Work}
\label{sec:future}

In future work, we would like to improve our use of the CLIP embeddings (e.g. by increasing our framerate) or use better, modality-matched embeddings to encode the video (e.g. using video-text pretrained models such as VideoCLIP~\citep{Xu2021VideoCLIPCP}). We would also like to dive deeper into the choice of shot transition detector; there is a rich body of literature with many different techniques for handling videos in different domains and settings~\citep{cotsaces2006video}. Finally, it would be interesting to improve our finetuning results and investigate how to adapt pretrained models in a sample-efficient manner for VMR.

\section{Conclusion}
We proposed a three-step zero-shot approach for Video Moment Retrieval and demonstrated how we vastly improve zero-shot performance, reducing the gap between zero-shot and state-of-the-art supervised models by a large margin. Furthermore, we also showed that our zero-shot approach beats non-pretrained supervised models on some metrics and comes close on other metrics and performs better than the best pretrained supervised model on shorter moments. We also ablated our hyperparameters and design choices and proposed interesting future directions.

\begin{ack}
We thank the students in the UT Austin CS398T ``Grounded Natural Language Processing'' course for their helpful feedback. 
\end{ack}

\bibliography{anthology, custom}
\bibliographystyle{acl_natbib}

\appendix
\section[Ablation for the lambda hyperparameter]{Ablation for the $\lambda$ hyperparameter}
\label{sec:applambda}
\begin{figure*}[h]
    \centering
    \includegraphics[width=1\textwidth]{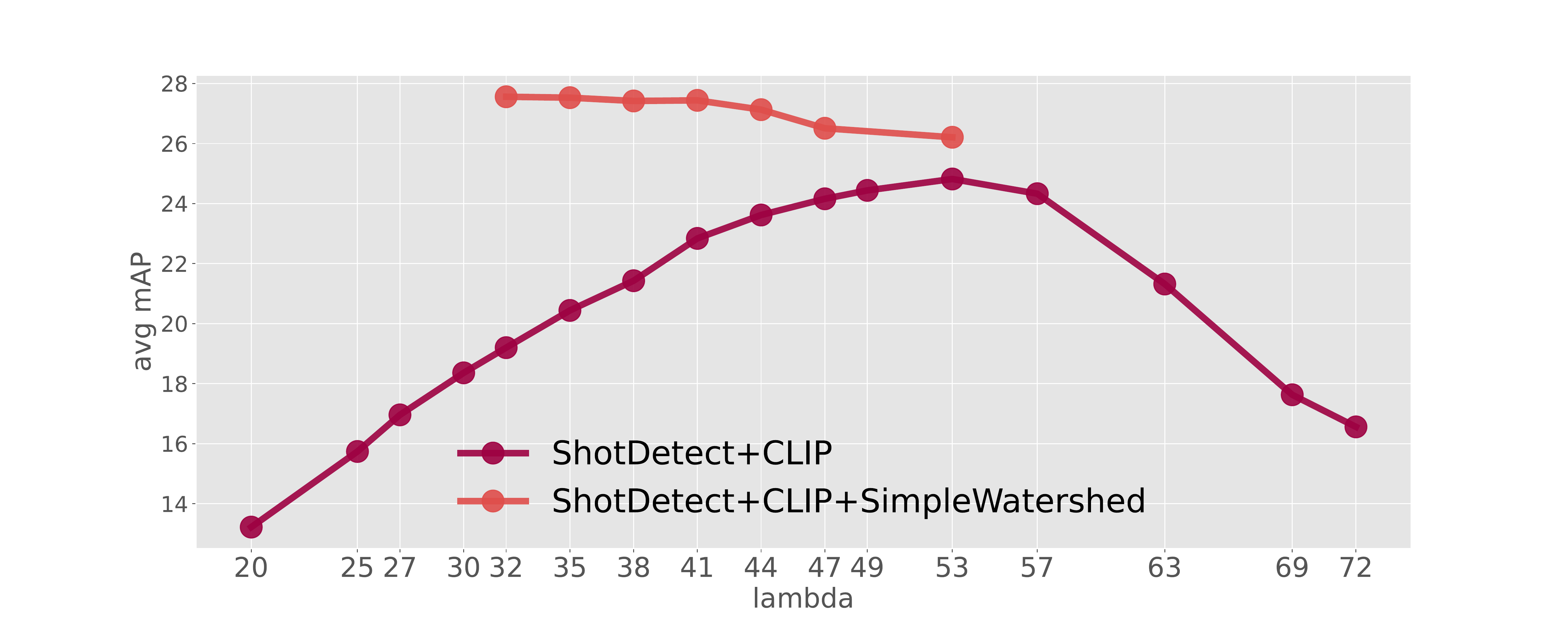}
    \caption{Ablating the threshold $\lambda$ hyperparameter} 
    \label{fig:thres}
\end{figure*}

We show the effect of the PySceneDetect threshold hyperparameter $\lambda$ by computing the avg mAP with for both ShotDetect+CLIP and ShotDetect+CLIP+SimpleWatershed (with the best postprocessing hyperparameter $\gamma$ pre-selected) in Figure~\ref{fig:thres}, showing how we obtained optimal $\lambda$ values of $53$ and $32$ for the two settings.

\section{The QVHighlights dataset.}
\label{sec:appendixqv}
\begin{figure}[h]
    \centering
    \vspace{-12pt}
    \includegraphics[width=1.\linewidth]{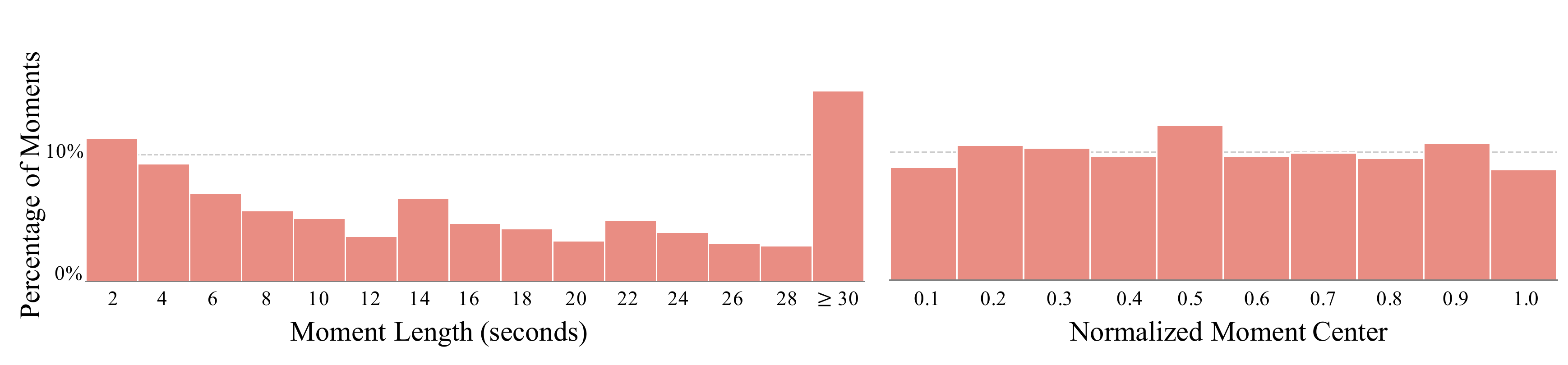}
    \caption{Distribution of moment lengths (\textit{left}) and normalized (by video duration) center timestamps (\textit{right}). The moments vary greatly in length, and they distribute almost evenly along the videos. Borrowed from~\cite{qvhighlights2021}
    } 
    \label{fig:moment_analysis}
\end{figure}

This dataset contains videos from user-created vlogs (categorized into `Daily Vlog' and `Travel Vlog') and news videos that have lots of raw footage. See Table~\ref{tab:stat_by_category} for video statistics by categories, and Figure~\ref{fig:moment_analysis} for statistics of ground truth moment. Both the figures and tables are borrowed from the QVHighlights paper~\citep{qvhighlights2021}.
\begin{table}[h]
    \small
    \centering
    \caption{Top unique verbs and nouns in queries, in each video category. Borrowed from~~\cite{qvhighlights2021}
    }
    \scalebox{1}{
        \begin{tabular}{lcll}
            \toprule
Category & \#Queries & Top Unique Verbs & Top Unique Nouns \\
\midrule
Daily Vlog & 4,473 & cook, apply, cut, clean & dog, kitchen, baby, floor \\
Travel Vlog & 4,694 & swim, visit, order, travel & beach, hotel, tour, plane \\
News & 1,143 & report, gather, protest, discuss & news, interview, weather, police \\
            \bottomrule
        \end{tabular}
    }
    \label{tab:stat_by_category}
    \end{table}

\end{document}